%% file: entry.tex
\documentclass[acmtog]{acmart}

\AtBeginDocument{%
  }

\setcopyright{None}
\copyrightyear{2024}
\acmYear{2024}
\acmDOI{XXXXXXX.XXXXXXX}

\acmSubmissionID{1631}
\acmJournal{TOG}
\acmVolume{11}
\acmNumber{2}
\acmArticle{33}
\acmMonth{10}

\usepackage{booktabs} 
\usepackage{threeparttable}
\usepackage{multirow}
\usepackage{tabularx}
\usepackage{amsfonts}
\usepackage{enumitem}
\usepackage{amsmath}
\usepackage{scalerel}
\usepackage{soul}
\usepackage{makecell}
\usepackage{xcolor} 
\usepackage{hyperref}
\usepackage{bbding}
\usepackage{booktabs}
\usepackage{colortbl}
\usepackage{xcolor}
\definecolor{best}{HTML}{FFCCCC}   
\definecolor{second}{HTML}{FFE5CC} 
\definecolor{third}{HTML}{FFFFCC}  

\newcommand{\first}[1]{\cellcolor{best}#1}
\newcommand{\second}[1]{\cellcolor{second}#1}
\newcommand{\third}[1]{\cellcolor{third}#1}
\definecolor{best}{HTML}{FFC7CE}    
\definecolor{second}{HTML}{FFE2C6}  
\definecolor{third}{HTML}{FFFACD}   

\newcommand{\name}{GeRM}

\newcommand{\Skip}[1]{}

\newcommand{\refSec}[1]{Section \ref{#1}}
\newcommand{\refFig}[1]{Fig. \ref{#1}}
\newcommand{\refTab}[1]{Table. \ref{#1}}

\newcolumntype{C}[1]{>{\centering\arraybackslash}m{#1}}

\newcolumntype{R}[1]{>{\raggedleft\arraybackslash}p{#1}}

\newcolumntype{L}[1]{>{\raggedright\arraybackslash}p{#1}}

\begin{document}

\title{GeRM: A Generative Rendering Model From Physically Realistic to Photorealistic}

\author {Jiayuan Lu}
\authornote{Both authors contributed equally to this research.}
\email{12321082@zju.edu.cn}
\affiliation{
 \institution{State Key Lab of CAD\&CG, Zhejiang University and Zhejiang Lab}
 \country{China}
}

\author {Rengan Xie}
\authornotemark[1]
\email{rgxie@zju.edu.cn}
\affiliation{
 \institution{State Key Lab of CAD\&CG, Zhejiang University and Zhejiang Lab}
 \country{China}
}

\author {Xuancheng Jin}
\email{xuanchengjin0@gmail.com}
\affiliation{
 \institution{State Key Lab of CAD\&CG, Zhejiang University and Zhejiang Lab}
 \country{China}
}

\author {Zhizhen Wu}
\email{zhizhenwu@zju.edu.cn}
\affiliation{
 \institution{State Key Lab of CAD\&CG, Zhejiang University and Zhejiang Lab}
 \country{China}
}

\author {Qi Ye}
\email{qi.ye@zju.edu.cn}
\affiliation{
 \institution{State Key Laboratory of Industrial Control Technology, Zhejiang University}
 \country{China}
}

\author {Tian Xie}
\email{rickyskyxie@zju.edu.cn}
\affiliation{
 \institution{Zhejiang University}
 \country{China}
}

\author {Hujun Bao}
\email{bao@cad.zju.edu.cn}
\affiliation{
 \institution{State Key Lab of CAD\&CG, Zhejiang University and Zhejiang Lab}
 \country{China}
}

\author {Rui Wang}
\email{ruiwang@zju.edu.cn}
\affiliation{
 \institution{State Key Lab of CAD\&CG, Zhejiang University and Zhejiang Lab}
 \country{China}
}

\author {Yuchi Huo}
\authornote{Corresponding author}
\email{huo.yuchi.sc@gmail.com}
\affiliation{
 \institution{State Key Lab of CAD\&CG, Zhejiang University and Zhejiang Lab}
 \country{China}
}

\renewcommand{\shortauthors}{Lu et al.}


\begin{teaserfigure}
    \includegraphics[width=1.0\linewidth]{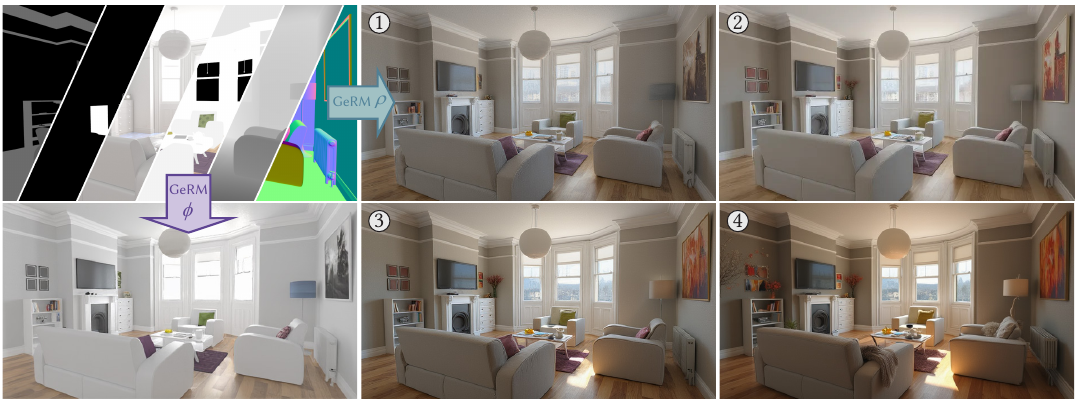}
    \vspace{-20pt}
      \caption{Given the physical attributes as input, the physically realistic ($\phi$) results produced by \name~are shown on the left side of the figure, with greater photorealism ($\rho$) as they move closer to the right. The sequence 1--4 illustrates this progressive realistic evolution: the model iteratively enriches the scene with realistic lighting dynamics (e.g., the warm sunlight and ground highlight), fine-grained material textures , and context-aware details (e.g., natural fabric wrinkles on the sofa), resulting in a high-fidelity photorealisitic image that consistent with the original image.
      }
  \label{fig:teaser}
\end{teaserfigure}

\input{section/0_abstract}



\maketitle

\input{samplebody-journals}
\end{document}

%% file: section/0_abstract.tex
\begin{abstract}
While physically-based rendering (PBR) simulates light transport that guarantees physical realism, achieving true photorealistic rendering (PRR) demands prohibitive time and labor, and still struggles to capture the intractable richness of the real world. We propose GeRM, the first multimodal generative rendering model to bridge the gap from PBR to PRR (P2P). We formulate this P2P transition by learning a distribution transfer vector (DTV) field to direct the generative process. To achieve this, we introduce a multi-condition ControlNet that synthesizes PBR images and progressively transitions them into PRR images, guided by G-buffers, text prompts, and cues for enhanced regions. To improve the model's grasp of the image distribution shift driven by text prompts, we propose a residual perceptual transfer mechanism to associate text prompts with corresponding targeted modification regions, which more clearly defines the incremental component updates. To supervise this transfer process, we introduce a multi-agent visual language model framework to construct an expert-guided pairwise transfer dataset, named P2P-50K, where each paired sample corresponds to a specific transfer vector in the DTV field. Extensive experiments demonstrate that GeRM synthesizes high-quality controllable images and outperforms state-of-the-art baselines across diverse applications, including PBR and PRR image synthesis and editing.

\end{abstract}

%% file: samplebody-journals.tex
\input{section/1_introduction}

\input{section/2_related_work}

\input{section/4_method}

\input{section/5_experiments}
\input{section/8_limitation}
\input{section/6_conclusion}
\newpage
\bibliographystyle{ACM-Reference-Format}
\bibliography{section/reference}

%% file: section/1_introduction.tex
\section{Introduction}
\label{sec:intro}
From the inception of computer graphics, one of the primary objectives has been clear and intuitive: to generate images that look exactly like photographs captured by a real camera~\cite{cook1982reflectance, ngan2005experimental}. For decades, the industry has focused on erasing the visual boundary between a computer-generated image and a realistic snapshot. Driven by this goal, modern physically based rendering (PBR) pipelines rely on rigorous simulation of physical laws. Built upon the rendering equation~\cite{kajiya1986rendering}, these pipelines model light transport and energy conservation to synthesize images. However, achieving realism through these simulations demands prohibitive human labor, time, and financial costs to manually tune rich micro-details and authentic lighting nuances. Constrained by these severe practical limitations, standard PBR outputs invariably compromise on intricate details and organic flaws. Consequently, they exhibit a persistent domain gap relative to actual photographs, leaving an evident boundary between computer-generated images and real-world photos.

\begin{figure}[htb]
    \centering
    \includegraphics[width=1\linewidth]{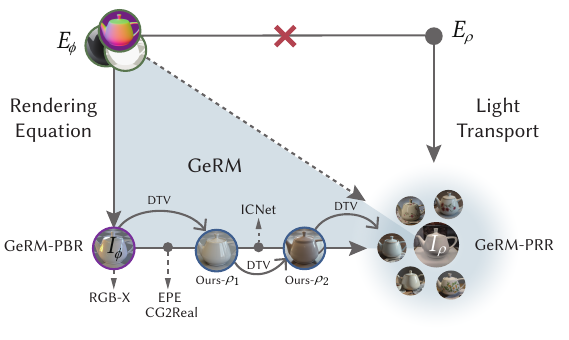}
    \caption{A \emph{P2P Quad} characterizing the correlation between physical realism and photorealsm.~$I_\phi$, $I_\rho$, $E_\phi$ and $E_\rho$ are physically realistic image, photorealistic image, digital existence, and real-world existence, respectively.}
    \label{fig:story}
\vspace{-0.3cm}
\end{figure}

To systematically analyze this boundary, we characterize the mappings between underlying existences and their corresponding visual observations as a \emph{P2P Quad} in Fig.~\ref{fig:story}. Within this quad, the photorealism of images synthesized by PBR pipelines is constrained by the digital essence ($E_\phi$). $E_\phi$ is a collection of digital parametric representations of the real world, including geometry, textures, and appearance models, intended to approximate the underlying real-world essence ($E_\rho$) as comprehensively as possible. Nonetheless, such an approximation is constrained by current physical models, limitations of acquisition hardware, computational capability, and the enduring quest to understand nature and realism, thereby implying a persistent gap between digital existence and real-world existence, such as those intractably rich and unenumerable detail. Therefore, PBR images ($I_\phi$) synthesized by the rendering equation, fall within a distribution distinct from PRR images ($I_\rho$), which is light transport result of real-world existence. How to synthesize a target photorealistic image from a digital existence 
remains an open problem in current human knowledge, as indicated by the dashed line in~\refFig{fig:story}. 

To bridge this gap, generative models~\cite{rombach2022high, goodfellow2014generative} leverage learned real-world distributions to naturally synthesize the organic and photorealistic details missing from computer-generated images. Although recent works have pioneered the integration of these generative approaches into rendering pipelines~\cite{richter2021enhancingphotorealismenhancement, cg2real, zeng2024rgb, Lu_2025_ICCV}, they share a critical bottleneck: their models are primarily trained on a single, unaligned distribution rather than explicit cross-domain trajectories. Forced to learn from this unconstrained singular distribution, they tend to overfit an averaged mapping, which inevitably leads to unstable realism or collapses into simple tone-mapping adjustments.

These limitations reveal three critical insights for bridging the P2P gap. \textit{First}, powerful generative priors are essential to compensate for simplified digital representations; strategically leveraging these priors enables us to construct high-quality paired data for cross-domain translation. \textit{Second}, because generative priors are inherently weakly constrained, they must be anchored by physical inputs (e.g., G-buffer signals) to prevent severe structural drift, ensuring that the photorealistic output preserves the foundational spatial layout of the original 3D scene while allowing for reasonable textural enrichment. \textit{Third}, overcoming the aforementioned instability of averaged mapping requires explicitly constructed paired P2P training data to teach the model a clear distribution transfer direction.

Building on these insights, we propose GeRM, a multimodal generative rendering model that unifies PBR and PRR synthesis. GeRM enables users to freely balance physical consistency and perceptual photorealism. We formulate the transition from PBR to PRR images as an iterative process, guided by a learned distribution transfer vector (DTV) field. To parameterize this field, we introduce a multi-condition ControlNet that synthesizes base PBR images and progressively injects photorealistic details into them, guided by G-buffers, text prompts, and localized cues. Furthermore, to effectively translate abstract text prompts into precise spatial updates, we propose a residual perceptual transfer mechanism that explicitly associates semantic intents with targeted modification regions, thereby clearly defining the incremental component updates at each step. Subsequently, to prevent over-editing and verify convergence during inference, we introduce an adaptive trajectory termination strategy based on semantic residual monitoring. Finally, to provide supervision for learning this transfer direction, we propose a multi-agent visual language model (VLM) framework to construct an expert-guided pairwise transfer dataset, named P2P-50K. 

Overall, our contributions are summarized as follows:
\begin{itemize}
 \item We propose GeRM, a multimodal generative rendering model that unifies PBR and PRR synthesis, enabling users to freely balance physical consistency and generative realism. It introduces a novel pathway that transforms idealized 3D scene representations into authentic PRR images.
 \item We design a multi-condition ControlNet to learn the DTV field, acting as a progressive neural renderer that effectively bridges strict physical G-buffer constraints with photorealistic semantic guidance.
 \item We propose a residual perceptual transfer mechanism that dynamically aligns abstract text prompts with specific spatial regions, ensuring accurate and localized photorealistic updates.
 \item We introduce an adaptive trajectory termination strategy via semantic residual monitoring, guaranteeing stable convergence during progressive rendering.
 \item We construct P2P-50K, a large-scale, expert-guided pairwise dataset curated by a multi-agent VLM framework, which provides explicit and robust supervision for the distribution transfer process.
\end{itemize}

%% file: section/2_related_work.tex
\section{Related Work}
\label{sec:relatedwork}

\subsection{Physically Realistic Rendering}
Physically realistic rendering aims to synthesize images that are consistent with the physical laws of light transport and surface–material interactions, emphasizing physically grounded and explainable image formation instead of purely perceptual realism.
This objective is formally characterized by the rendering equation \cite{kajiya1986rendering}, which models outgoing radiance as an integral of incoming illumination modulated by material reflectance and geometric visibility, providing a unified theoretical foundation for physically based image rendering.
Building upon this formulation, offline rendering techniques such as Monte Carlo path tracing \cite{veach1998robust} and global illumination \cite{JensenGI, LTSV} explicitly simulate multi-bounce light transport and complex indirect lighting effects, and are widely regarded as the gold standard for photorealistic rendering despite their substantial computational cost.
In addition to accurate light transport, achieving high visual realism critically depends on faithful scene modeling, including physically based material representations \cite{cook1982reflectance, WalterMicrofacet} and specialized appearance models for complex surfaces such as hair fibers \cite{AutoHair} and woven cloth \cite{RealCloth}, which cannot be adequately captured by simple reflectance assumptions.
However, due to the inherent complexity of real-world illumination, geometry, and material variability, faithfully reproducing photorealistic appearance under diverse conditions remains a challenging and active research problem.

\begin{figure*}[h]
    \centering
    \includegraphics[width=1\linewidth]{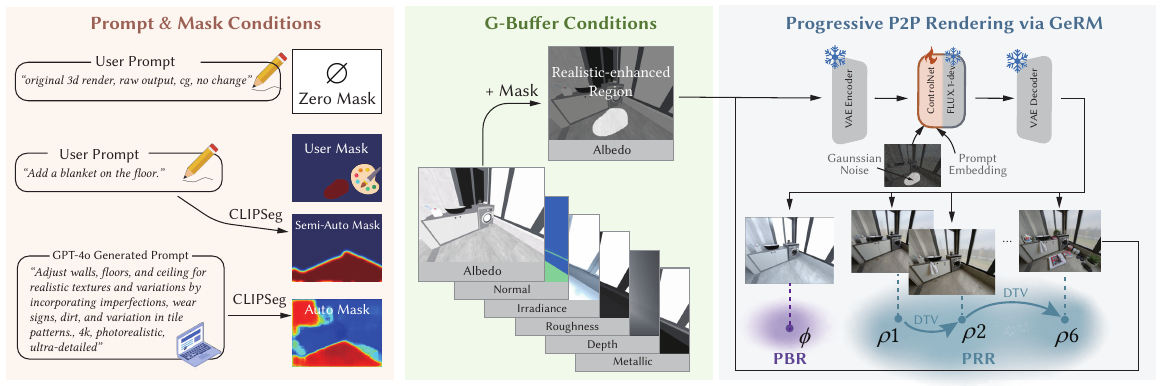}
    \caption{Overview of our GeRM framework. 
Our pipeline operates on a multi-condition framework that integrates physical G-buffers with task-adaptive spatial constraints—ranging from zero masks for PBR synthesis to user-specified regions or VLM-driven automatic maps for editing and progressive realism. 
These combined inputs are fed into the multi-condition ControlNet, which predicts a distribution transfer vector (DTV) field. 
This DTV field guides the iterative evolution of the scene, transforming the initial physical state ($\phi$) into the final photorealistic target ($\rho_6$) via iteratively progressive generation.}
    \label{fig:overview}
\vspace{-0.3cm}
\end{figure*}

\subsection{Generative Models for Image Synthesis}
Recent years have witnessed significant progress in generative models for image synthesis, with diffusion-based \cite{dhariwal2021diffusion, sdxl} and flow-based methods \cite{flowMatching, rectifiedFlow, flowStoINter} emerging as two dominant paradigms.
These approaches model complex image distributions through iterative stochastic or deterministic processes, enabling high-quality image generation and distributional control.
Owing to their strong expressive power and generative capability, recent generative models have motivated growing interest in extending their use beyond conventional image synthesis, including their integration into neural rendering pipelines \cite{zeng2024rgb, DiffusionRenderer}.
Many of these approaches can be interpreted as learning transformations between probability distributions through iterative stochastic or deterministic processes \cite{flowIntro}.
In particular, flow matching formulates generation as the integration of a learned vector field that simulates an ordinary differential equation (ODE), providing a efficient framework for transforming simple source distributions into complex target distributions.
Inspired by this perspective, our work adopts a distributional transformation viewpoint for rendering: instead of generating images from scratch, we formulate rendering as a process that progressively transforms physically based renderings toward photorealistic appearance, while preserving physical consistency through the consistent injection of G-buffer representations as conditioning signals.

\subsection{Generative Rendering and Photorealism Enhancement}
Achieving true photographic realism (PRR) through traditional Physically-Based Rendering (PBR) is notoriously resource-intensive. It demands immense human labor, time, and financial costs to meticulously design high-fidelity assets and manually tune complex lighting, textures, and micro-details. To circumvent these expenses, recent efforts leverage large-scale generative models. By utilizing real-world visual priors, these models can naturally synthesize physical details, drastically reducing rendering overhead. For instance, RGB$\leftrightarrow$X~\cite{zeng2024rgb} and Nvidia Cosmos~\cite{DiffusionRenderer} utilize diffusion models for G-buffer-conditioned synthesis. Concurrently, methods like EPE~\cite{richter2021enhancingphotorealismenhancement} and IntrinsicControlNet (ICNet)~\cite{Lu_2025_ICCV} attempt to directly enhance photorealism from physical buffers. However, lacking paired cross-domain training data, these methods rely on idealized synthetic datasets or unpaired photos. Without explicit transitional guidance, they often collapse into simple tone-mapping adjustments or yield unstable structural realism. Recognizing this bottleneck, we formulate photorealism enhancement as a controllable distribution transfer problem, learning a definitive translation direction via a novel expert-guided dataset (P2P-50K). However, the unconstrained image editing model used to construct this dataset is unsuitable for direct inference due to its weak physical control and low success rate. Instead, by distilling this curated data into our framework, we achieve an efficient pipeline that seamlessly bridges the PBR-to-PRR spectrum while maintaining strict physical consistency through explicit G-buffer conditioning.

%% file: section/4_method.tex
\section{Method}
\label{sec:pro_state}
To address the challenge of bridging the gap between PBR and PRR, we propose a progressive G-buffer and prompt-guided generative rendering framework. 
We formulate photorealistic rendering as an iterative distribution transfer process (\refSec{sec:3-1-statement}), aiming to learn a vector field that evolves PBR distribution into PRR distribution. 
To parameterize the distribution transfer vector (DTV) field, we develop a multi-condition ControlNet (\refSec{sec:3-3-controlnet}). This architecture employs a unified conditioning strategy to predict the precise state update, enabling both PBR synthesis and progressive PRR generation.
Furthermore, to bridge the perceptual gap between textual prompts and spatial updates for controllable incremental realism injection, we introduce a residual perceptual transfer mechanism (\refSec{sec:3-4-clipseg}) that projects semantic intentions into explicit spatial attention fields.
Subsequently,  to verify the convergence of the iterative generation during inference, we implement an adaptive trajectory termination strategy (\refSec{sec:3-6-converge}) based on semantic residual monitoring. Finally, to compensate for the absence of explicit learnable metrics, we construct an expert-guided progressive pairwise P2P dataset (\refSec{sec:3-2-dataset}), which establishes the refinement trajectory through iterative steps guided by a multi-agent VLM.

\subsection{Problem Statement}
\label{sec:3-1-statement}

From a generative modeling perspective, our goal is to train a model that produces samples from a target distribution conditioned on a set of control signals, which can be generally formulated as conditional generation.
Such problems are commonly addressed by conditional flow-based or diffusion models.
When paired training data are available, models can directly learn a conditional distribution $p_\theta(\mathbf{x} \mid \mathbf{C})$.
Alternatively, generation can be guided by reweighting a prior distribution with an energy function $\ell(\mathbf{C}, \mathbf{x})$, yielding samples drawn from:
\begin{equation}
\mathbf{x} \sim \exp\!\big[-\ell(\mathbf{C}, \mathbf{x})\big]\cdot p_\theta(\mathbf{x}),
\end{equation}
where the energy term measures the alignment between the synthesized image $\mathbf{x}$ and the conditioning signal $\mathbf{C}$, which in our case $\mathbf{x}$ denotes the rendered image to be synthesized and $\mathbf{C}$ represents physically based material properties together with other rendering-related control signals.

However, formulating photorealistic rendering as a standard conditional generation problem poses fundamental challenges.
First, the absence of paired PBR--PRR image data prevents directly learning the conditional distribution $p_\theta(\mathbf{x} \mid \mathbf{C})$.
Second, the lack of an explicit, optimizable supervision signal for perceptual realism makes it infeasible to define a meaningful energy function $\ell(\mathbf{C}, \mathbf{x})$.
Consequently, neither direct conditional learning nor energy-based guidance paradigms are applicable in our setting.

Motivated by these limitations, we avoid directly modeling the direct PBR-to-PRR conditional distribution.
Instead, we reformulate photorealistic rendering as a progressive transition process, where the generation is decomposed into multiple incremental steps.
At each step, the model is tasked with predicting only distribution transfer vector (DTV) that moves the current rendering toward increased photorealism under the given control conditions.
Concretely, we learn a conditional distribution
\begin{equation}
    p_\theta(\Delta \mathbf{x} \mid \mathbf{x}, \mathbf{C}),
\label{eq:state_vector}
\end{equation}
where $\Delta \mathbf{x}$ represents a DTV applied to the current image.
Under this formulation, the original challenges of defining a realism metrics or learning a direct PBR--PRR mapping are transformed into the more tractable problem of constructing paired datasets between a PBR rendering and a relatively more photorealistic intermediate result, enabling effective learning of the distribution transfer vector field,
\begin{equation}
  \mathbf{f}_\theta( \mathbf{x'}\mid \mathbf{x}, \mathbf{C}),
\label{eq:state_vector_field}
\end{equation}
where $\mathbf{x'}=\mathbf{x}+\Delta \mathbf{x}$, $\mathbf{x'}$ is the target under the distribution transfer from $\mathbf{x}$.

\subsection{Multi-Condition ControlNet for PBR Consistent Image Synthesis and Transfer}
\label{sec:3-3-controlnet}
To learn the conditional distribution of the DTV field defined in Eq.~\eqref{eq:state_vector_field}, we design a multi-condition ControlNet optimized on our constructed paired P2P-50K dataset. 
During inference, the process begins by synthesizing a base PBR image derived from the G‑buffer and the text prompt. 
Subsequently, this representation is iteratively refined toward the PRR target by augmenting the physical anchors with semantic modulation. 
Within this framework, the current rendering $\mathbf{x}_t$ acts as a realistic prior which leverages the integrated physical and semantic controls to direct the network to generate the $\mathbf{x}_{t+1}$, which transitions from
$\mathbf{x}_{t}$ and approaches PRR more closely.

Formally, we employ a unified conditioning strategy to synthesize from the PBR to the PRR distribution. 
In all stages, the network accepts a 21-channel control tensor constructed by concatenating the physical conditions $\mathbf{C} \in \mathbb{R}^{H \times W \times 18}$—comprising six PBR buffers (albedo, roughness, metallic, normal, depth, and irradiance)—with a semantic modulation mask $\mathbf{M}_t \in \mathbb{R}^{H \times W \times 3}$, which serves as a spatial attention field derived from the cross-modal alignment mechanism (detailed in \refSec{sec:3-4-clipseg}).

Consequently, the network predicts the more PRR $\mathbf{x}_{t+1}$ at time step $t$ as follows:
\vspace{-5pt}
\begin{equation}
    \mathbf{x}_{t+1} = \mathbf{f}_\theta(\mathbf{z}_t, t, \mathbf{C}, \mathbf{M_t}, \mathbf{y}_t),
\label{eq:flow_pred}
\vspace{-0.5pt}
\end{equation}
where the $\mathbf{f}_\theta$ is our multi-condition ControlNet, regarded as the DTV field, with parameter $\theta$, $\mathbf{y}_t$ is the textual instruction, and $\mathbf{z}$ is a noisy latent of $\mathbf{x}_t$ for consistent transition: for the initial PBR synthesis, it is sampled from a standard Gaussian distribution $\mathbf{z}_0 \sim \mathcal{N}(\mathbf{0}, \mathbf{I})$, whereas for the PRR evolution, it is initialized from the encoding of the previous state
$\mathbf{z}_t = \mathcal{E}(\mathbf{x}_t)$,
where $\mathcal{E}(\cdot)$ denotes an encoder that maps the current rendering state into a latent representation.
Also we define the DTV $\Delta {\mathbf{x}_t}$ as:
\begin{equation}
    \Delta \mathbf{x}_{t} = 
    \begin{cases} 
        \mathbf{f}_{\theta}(\mathbf{z}_t, t, \mathbf{C}), & \text{$t=0$}, \\
        \mathbf{f}_{\theta}(\mathbf{z}_{t}, t, \mathbf{C}, \mathbf{y}_{t})
        -
        \mathbf{f}_{\theta}(\mathbf{z}_{t-1}, t-1, \mathbf{C}, \mathbf{M}_{t-1}, \mathbf{y}_{t-1}), & \text{$t>0$} .
    \end{cases}
\label{eq:switch}
\end{equation}
Note that the continuous injection of G-buffer conditions $\mathbf{C}$ acts as a physical anchor, actively rectifying the spatial drift and error accumulation that typically plague standard iterative editing frameworks.

Finally, to learn the underlying distribution transfer vector field that dictates the state updates $\Delta \mathbf{x}_{t}$ under unified controls spanning both PBR initialization and progressive PRR evolution, we train the network to learn the DTV from the constructed paired dataset, which provides supervision on how the rendering state should evolve toward photorealism.
This learning process can be formulated as follows:
\begin{equation}
\mathcal{L} = \mathbb{E}_{t, x_{t+1}, x_t} \left[ \Delta \hat{x}_t - (x_{t+1} - x_t) \right]_{2}^{2} ,
\label{eq:loss}
\end{equation}
where $t$ represents the step, $\mathbf{x}_t$ denotes the source state originating from either Gaussian noise or the encoded state of the previous iteration, $\mathbf{x}_{t+1}$ corresponds to the target ground truth from our paired P2P dataset and $\Delta \hat{x}_t$ is the final DTV predicted by our model. Please see \textit{Appendix C.1} for training details.

To further enhance the flexibility of this physical control, we enable the network to robustly handle partial inputs and foster physical channel decoupling, allowing for the independent editing of specific G-buffer subsets. 
To this end, we employ a Bernoulli channel dropout strategy during training
\vspace{-1.5pt}
\begin{equation}
    \mathbf{C} = \text{Concat}(\mathbf{c}_1 \cdot b_1, \dots, \mathbf{c}_6 \cdot b_6), \quad \text{where } b_i \sim \text{Bernoulli}(p_i),
\label{eq:dropout}
\end{equation}
\vspace{-1.5pt}
where $\mathbf{c}_i$ ($1 \le i \le 6$) represents the specific G-buffer component corresponding to albedo, roughness, metallic, normal, depth, and irradiance, respectively, and $b_i \in \{0, 1\}$ indicates the binary inclusion state for each buffer sampled with retention probability $p_i$.

\subsection{Enhance Perception of Image Transition Increment}
\label{sec:3-4-clipseg}

As established in Eq.~\eqref{eq:flow_pred}, we formulate the P2P process as a distribution transfer, where each iteration $\mathbf{x}_t \to \mathbf{x}_{t+1}$ represents an incremental shift along a trajectory towards the target distribution. 
Ideally, this evolution should be perceptibly guided by the textual instructions $\mathbf{y}$, where specific semantic changes in the prompt dictate the corresponding spatial updates in the image.

However, a critical perceptual gap arises in this multi-modal mapping is that the network often struggles to establish a correspondence between the textual description of change and the actual visual distribution transition. 
Even when trained on our pairwise P2P-50K dataset, the model lacks explicit spatial guidance to associate specific semantic tokens (e.g., ``add cracks'' or ``remove clutter'') with their resulting significant visual differentials. 
Consequently, the model fails to perceive which part of the prompt is responsible for driving the distribution shift, leading to a failure where the predicted $\mathbf{v}_\theta$ lacks spatial focus, resulting in weak or irrelevant updates that ignore fine-grained instructions. 

To bridge this disconnect and effectively align the semantic differential with the spatial incremental update, we introduce residual perceptual transfer mechanism for cross-modal alignment that employs a mapping strategy to directly translate abstract semantic instructions $\mathbf{y}$ into an explicit attention mask $\mathbf{M}$.
First, we extract the target entity from the textual prompt and project it onto the current PRR visual state $\mathbf{x}_t$:
\vspace{-3pt}
\begin{equation}
    e_{1} = SpaCy(\mathbf{y_t}), \quad \mathbf{M}_{raw} = \mathcal{S}_{clip}(\mathbf{x}_t, e_{1}).
\label{eq:clipseg_base}
\end{equation}
Specifically, $\mathbf{M}_{raw}$ is determined through a flexible dual-source strategy. In automatic mode, we employ the linguistic parser SpaCy~\cite{spacy2} to analyze the syntactic structure of the prompt $\mathbf{y}$, filtering out abstract command verbs to isolate the salient noun entity $e_{1}$. We then utilize the CLIPSeg encoder $\mathcal{S}_{clip}$~\cite{clipseg}, a zero-shot segmentation model capable of performing pixel-wise alignment between visual features and text queries, to generate a raw probability map $\mathbf{M}_{raw}$, which is further filtered by an adjustable empirical confidence threshold. Alternatively, for precise manual control, this map can be explicitly substituted with a user-provided region ($\mathbf{M}_{raw} = \mathbf{M}_{user}$).

To ensure the incremental update $\Delta \mathbf{x}_{pred}$ (as defined in Eq.~\eqref{eq:switch}) blends seamlessly with the surrounding context, we apply a morphological smoothing operation to $\mathbf{M}_{raw}$:
\begin{equation}
    \mathbf{M}_t = \left( \mathbf{M}_{raw} \oplus \mathcal{K}_{dil} \right) \ast \mathcal{G}_{\sigma}.
\label{eq:morph}
\end{equation}
Here, $\oplus$ denotes morphological dilation with a structural element $\mathcal{K}_{dil}$, and $\ast$ represents convolution with a Gaussian kernel $\mathcal{G}_{\sigma}$.
Regardless of the mask source (automatic or user-specified), this post-processing is critical for visual consistency. 
Specifically, the morphological dilation ensures the mask fully encompasses the target object to prevent boundary leakage. 
Simultaneously, the Gaussian smoothing creates a soft transition zone that forces the update magnitude to decay gradually at the edges, effectively preventing hard-seam artifacts and ensuring a natural fusion between the modified region and the background.


\subsection{Adaptive Trajectory Termination via Semantic Residual Monitoring}
\label{sec:3-6-converge}
Driven by the learned DTV field, our framework iteratively evolves the source PBR state toward the target PRR distribution during inference. However, the precise trajectory length required to fully achieve photorealism varies across different scenes and cannot be known in advance. Empirically, we observe a distinct convergence behavior (as visualized in Fig. ~\ref{fig:converge}): as the generated image gets close to the photorealistic state, the model stops making significant changes and instead starts to oscillate slightly around the final PRR result. 
Consequently, arbitrarily enforcing a fixed iteration count is rigid; we require a rational mechanism to verify convergence. 
Once the distribution transfer reaches the target PRR state, the distribution transfer vector field naturally diminishes, meaning that subsequent iterations yield negligible perceptual improvements.

To adaptively determine this stopping point, we introduce the semantic intensity ($I_t$) check. 
This metric integrates two critical indicators to monitor the progressive generation status: 
(1) the pixel-wise residual, where a decrease indicates that the generative backbone considers the current region sufficiently optimized and requires no further modification; 
and (2) the semantic attention map, where a diminishing response indicates that the VLM considers the image sufficiently photorealistic and ceases to propose actionable refinement prompts.
Mathematically, $I_t$ is calculated as the mask-weighted sum of the residuals between the current state and the predicted next state, ensuring that a decline in either factor triggers termination:
\begin{equation}
    I_t = \mathbf{M}_{t} \cdot \| \mathbf{\hat{x}}_{t+1} - \mathbf{x}_{t} \|_1 ,
\label{eq:convergence}
\end{equation}
Here, $\mathbf{M}_{t}$ ensures the metric focuses strictly on the semantic area of interest, while the residual term $\|\mathbf{\hat{x}}_{t+1} - \mathbf{x}_{t}\|_1$ detects the saturation of the generation process. 
When $I_t$ falls below a threshold $\tau_{stop}$, we consider the scene converged to the PRR distribution and terminate the inference. This termination strategy is empirically validated in our convergence experiments (see \refSec{sec:converge_ablation}, Fig.~\ref{fig:converge}, and Fig.~\ref{fig:converge_analysis}), which demonstrate that $I_t$ effectively signals the saturation point of our P2P transfer.

\begin{figure*}[htb]
    \centering
    \includegraphics[width=1\linewidth]{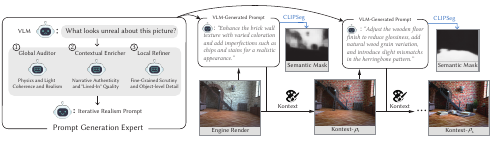}
    \caption{Pipeline for constructing the progressive pairwise P2P dataset. We employ a multi-agent VLM framework—comprising the \textit{Global Auditor}, \textit{Contextual Enricher}, and \textit{Local Refiner}—to collaboratively generate realistic refinement prompts based on the current state. 
These prompts are fed into CLIPSeg to produce spatial semantic masks, accurately localizing the regions for realistic modification. 
Guided by these spatially grounded instructions, the generative backbone (Kontext) executes localized updates, progressively evolving the sterile Engine Render into a photorealistic state (Kontext-$\rho_t$).}
    \label{fig:dataset}
\vspace{-0.3cm}
\end{figure*}

\begin{figure}[htb]
    \centering
    \includegraphics[width=1\linewidth]{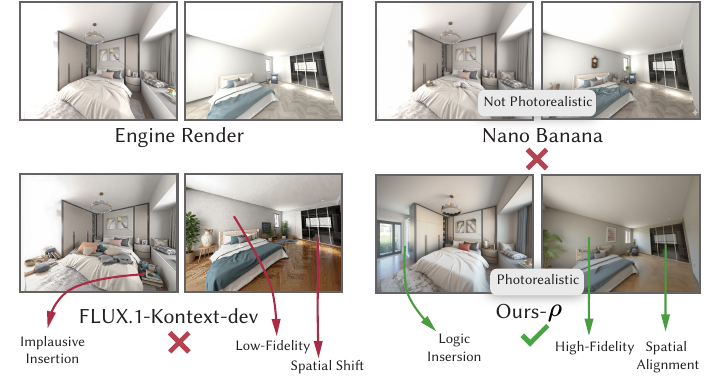}
    \caption{
    Why SOTA Editing Models Are Insufficient for Photorealistic Generation. 
    Under the progressive prompts shortened like ``physically realistic appearance,'' Nano Banana retains a synthetic appearance, while Kontext exhibits severe spatial instability. In contrast, our method achieves superior photorealism and spatial alignment.
}
    \label{fig:intro}
\vspace{-0.3cm}
\end{figure}

\subsection{Expert-Guided Progressive Pairwise P2P
 Transfer Dataset Construction}
\label{sec:3-2-dataset}
To effectively learn the conditional distribution $p_\theta$ defined in Eq.~\eqref{eq:state_vector_field}, we require high-quality training pairs that represent the trajectory from the source PBR state $\mathbf{x}_0$ to the PRR target $\mathbf{x}_T$. However, acquiring such aligned data is non-trivial. 
To bridge this gap, we construct the expert-guided progressive pairwise realism dataset, a large-scale collection designed to provide supervision signals for physically consistent, progressive photorealistic image generation.

To ensure comprehensive scene fidelity and reduce the semantic hallucinations common in single-pass descriptions, we adopt a collaborative approach. As shown in Fig.~\ref{fig:dataset}, we employ a multi-agent visual language model (VLM) framework to act as a prompt generation expert. 
Within this framework, three specialized agents collaborate to critique the scene: 
The \textit{Global Auditor} evaluates the overall image quality, specifically checking for unnatural lighting and overly simplified surfaces; 
the \textit{Contextual Enricher} upgrades the scene fidelity, compensates simple shapes and plain materials with realistic details; 
and the \textit{Local Refiner} refines the layout by adding reasonable items to empty areas to make the scene look more complete.
Based on this collective critique, the agents generate a textual refinement prompt, which is subsequently processed by CLIPSeg~\cite{clipseg} to produce a corresponding spatial mask $\mathbf{M}_t$.
This semantic mask localizes the update to specific regions, acting as a spatial guide for the DTV $\Delta \mathbf{x}_t$ from the PBR to the PRR distribution at time step $t$.

Conditioned on these instructions, we utilize FLUX.1-Kontext-dev as the backbone to execute the progressive distribution transfer $\mathbf{x}_t \to \mathbf{x}_{t+1}$.  
We prioritize this model for its exceptional photorealistic capability. However, as shown in Fig.~\ref{fig:intro}, lacking explicit G-buffer conditioning, it often introduces severe geometric or material inconsistencies. Conversely, alternatives like Nano Banana~\cite{nanobanana2025} provide better structural stability but fundamentally retain a flat, synthetic appearance. Weighing this trade-off, we choose to build upon the photorealistic model; we argue that rectifying its spatial and semantic artifacts via post-processing is significantly more tractable than attempting to induce true realism into a sterile output.

To this end, we implement a rigorous post-generation curation mechanism to ensure the dataset captures valid incremental changes. First, we apply spatial alignment via homography estimation to correct minor pixel shifts inherent to the unconstrained generative process, ensuring pixel-perfect consistency with the original G-buffers. Second, we employ a visual-semantic alignment strategy that calculates pixel-wise residuals masked by semantic intent. By discarding samples where the intended changes fail to manifest or exhibit excessive noise, we guarantee that the final optimization focuses exclusively on effectively transferred regions. 

This curated generation process operates dynamically: at each iteration step $t+1$, the multi-agent system re-evaluates the newly generated image $\mathbf{x}_{t}$ to propose a fresh, context-aware prompt for subsequent refinement. While increasing the iterations theoretically allows for a smoother transition from the PBR to the PRR distribution, excessive steps introduce considerable error accumulation. Based on our training trajectory length selection experiment, we identify six steps as the optimal trade-off between photorealism and stability. 
Comprehensive details regarding the length selection experiment, data curation strategies, and visualizations of the refined P2P dataset samples are provided in \textit{Appendix C.3, C.4, C.2}.

Ultimately, this pipeline yields a robust and diverse collection of 50,000 paired progressive P2P training samples. To ensure comprehensive coverage across different environments, this dataset comprises 35,000 indoor scenes sourced from Hypersim~\cite{hypersim} and \citet{kjl}, 5,000 outdoor scenes from MatrixCity~\cite{li2023matrixcity}, and 10,000 single-object assets from Objaverse~\cite{objaverse}.

\begin{figure*}[!h]
    \centering
    \includegraphics[width=1\linewidth]{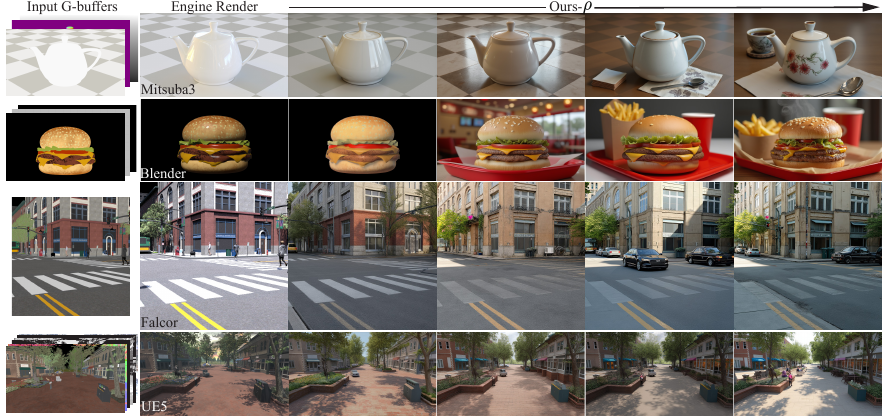}
    \vspace{-20pt}
    \caption{Visual comparison of our iterative PRR generation against standard rendering engines. From left to right: input G-buffers, the baseline render from various engines (Mitsuba3, Blender, Falcor, and UE5), and the PRR results generated by our method. As shown in the four right columns, our approach gradually enhances the realism of the scene, generating details and lighting effects that surpass the original engine renders.}
    \label{fig:comparison-engine}
    \vspace{-0.1cm}
\end{figure*}

\begin{figure*}[!h]
    \centering
    \includegraphics[width=1\linewidth]{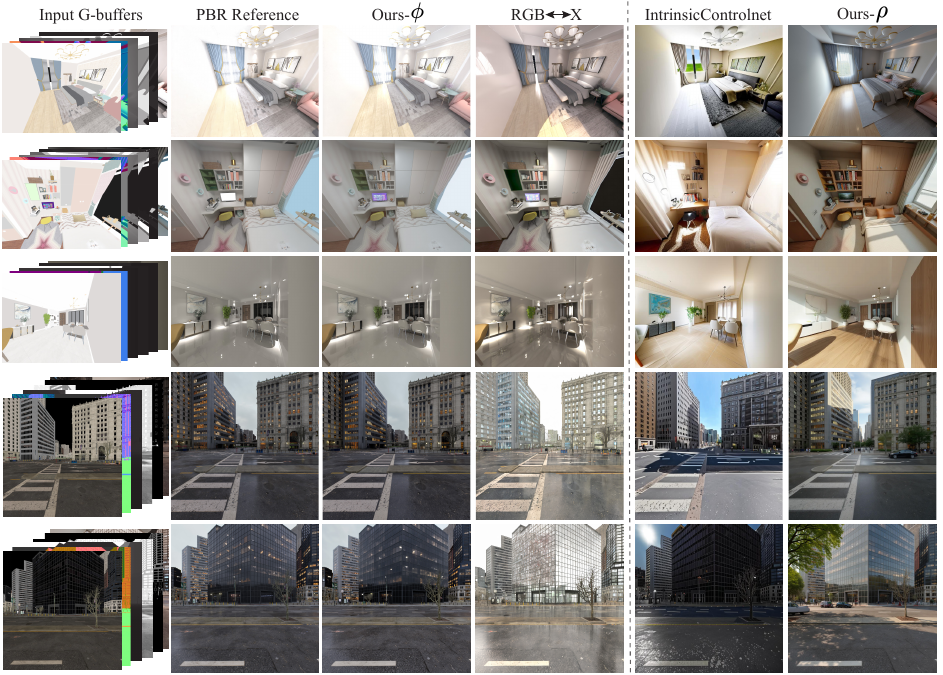}
    \caption{Comparisons of PBR and PRR synthesis on indoor and outdoor scenes. Given input G-buffers (leftmost), the region left of the dashed line validates physical accuracy against PBR Reference and RGB$\leftrightarrow$X. The region to the right evaluates our final PRR generation against IntrinsicControlNet. 
    }
    \label{fig:comparison-1}
    \vspace{-0.3cm}
\end{figure*}

%% file: section/5_experiments.tex
\begin{figure*}[!h]
    \centering
    \includegraphics[width=1\linewidth]{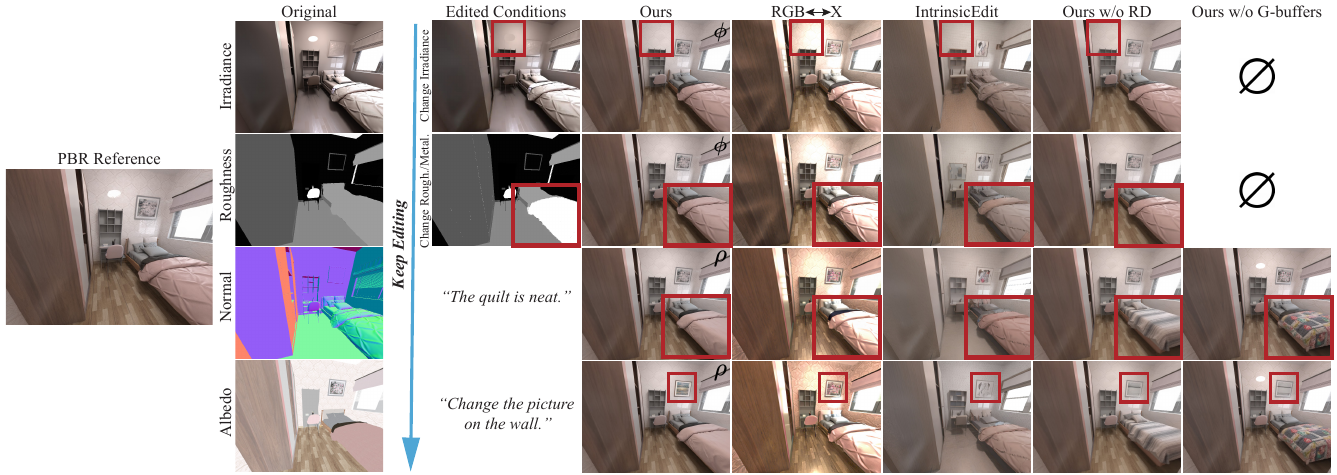}
    \vspace{-20pt}
    \caption{Demonstration of progressive editing on different subsets of G-buffer channels. We visualize an editing sequence where modifications are applied sequentially from top to bottom. In each row, the generation builds upon the result from the previous step, controlled by the specific edited G-buffer channel or text prompt shown in the ``Edited Conditions'' column. We compare our method against RGB$\leftrightarrow$X and IntrinsicEdit, alongside the ablation variants without random dropout (Ours w/o RD) or G-buffer guidance (Ours w/o G-buffers). Additional editing cases are provided in the \textit{Appendix A.4}.
}
    \label{fig:ablation_channel}
    \vspace{-0.3cm}
\end{figure*}


\begin{figure*}[!h]
    \centering
    \includegraphics[width=1\linewidth]{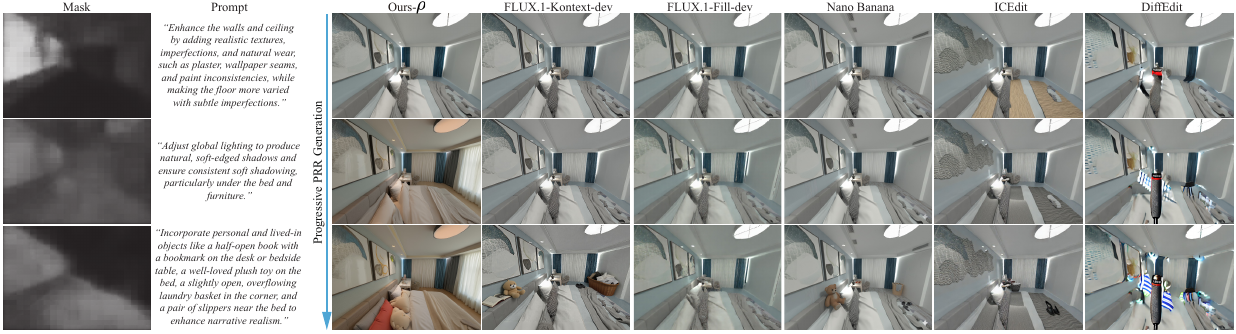}
    \vspace{-20pt}
    \caption{Progressive semantic-aware P2P generation. We visualize masks, prompts, and generation results across progressive steps. While existing baselines struggle to balance strict localized mask control with true photorealism, our approach successfully confines high-fidelity enhancements to designated areas, preserving unmasked regions.
    }
    \label{fig:mask_real}
\end{figure*}

\section{Experimental Setup}
\label{sec:experiments}
\subsection{Implementation Details}
We trained our model on our constructed pairwise P2P-50K dataset. 
All images were resized to 512 \(\times\) 512 pixels. The training was carried out on the FLUX.1-dev ControlNet \cite{flux2024} using two 80GB GPUs equivalent to the NVIDIA A100 with a batch size of 32. Importantly, to demonstrate true generalization, all the visual results shown here are generated from data not included in the training set.

\subsection{Evaluation Metrics}
Quantifying photorealism remains an open challenge in the community, with no tailored metric currently available, as mentioned in \refSec{sec:3-1-statement}. Consequently, we combine available quantitative metrics with user studies to comprehensively assess our method across different sub-tasks.

\subsubsection{Quantitative Metrics.}
To conduct a rigorous and objective evaluation of our framework, we categorize the employed metrics into two primary aspects:

To evaluate PBR synthesis accuracy and structural preservation, we utilize PSNR, SSIM~\cite{psnr&ssim}, and LPIPS~\cite{lpips} to measure pixel-level fidelity and perceptual sharpness against the source image, ensuring background preservation in unmasked regions. To enforce structural plausibility, we employ DINOv2 Similarity~\cite{oquab2023dinov2}, which leverages geometry sensitive features to ensure adherence to the semantic layout. Additionally, we use  CLIP (Global-I) score to measure the high-level visual semantic similarity between our PBR output and the reference PBR image, validating the accuracy of our PBR process.

To assess the generative PRR quality and cross-modal semantic alignment, we first we calculate the KID~\shortcite{kid} between 300 generated samples and the ADE20K~\cite{ade20k} dataset. Since ADE20K comprises diverse real-world photographs, this distributional distance can quantify how closely our PRR images approximate true realism, effectively measuring our success in bridging the P2P gap. 
For semantic alignment, we employ CLIP Score (Text)~\cite{clip} across three dimensions: \textit{Global} for overall image-text consistency, \textit{Local} for semantic alignment within the modified region, and \textit{Directional} to measures how well the visual changes applied to the original image correspond to the text instruction. We further utilize BLIP~\cite{li2022blip} to evaluate vision-language alignment, ensuring the generated content semantically matches the prompt descriptions. Finally, to quantify human-aligned quality, we rely on ImageReward (IR)~\cite{xu2023imagereward} and Aesthetic Score (Aes)~\cite{aes}, which serve as proxies for human preference and artistic value. 
Additionally, we report $\Delta$CLIP, defined as the net increase in the global CLIP score relative to the source image, to measure the absolute improvement in semantic alignment. 
To evaluate the comprehensive trade-off between structural preservation and editing accuracy, we introduce Q-Score, calculated as the harmonic mean of SSIM and Local CLIP score.

\begin{table}[h]
    \centering
    \caption{Quantitative evaluation of photorealism. US denotes user study.
    To measure the capability of generating photorealistic images and details, we employ highly detailed realistic prompts generated by GPT-4o. 
    Our method achieves state-of-the-art performance in terms of realism and text-image alignment.
    \colorbox{best}{Red}, \colorbox{second}{Orange}, and \colorbox{third}{Yellow} indicate top 3 results.}
    \vspace{-10pt}
    \begin{tabular*}{\linewidth}{l c c c c}
        \toprule
        & US $\uparrow$ & KID $\downarrow$ & CLIP (Global-T) $\uparrow$ & IR $\uparrow$ \\
        \midrule
        Ours-$\rho$3      & \second{3.52} & \second{0.05834} & \second{0.2883} & \third{-0.3197} \\
        Ours-$\rho$6                & \first{3.99}  & \first{0.04572}  & \first{0.2920}  & \first{-0.2084} \\
        ICNet  & 1.40          & 0.06804          & 0.2806          & -0.5022         \\
        FLUX.1-Kontext   & \third{2.50}  & \third{0.05975}  & 0.2830          & \second{-0.3017}\\
        Nano Banana          & 2.48          & 0.09024          & 0.2783          & -0.4468         \\
        Mitsuba3             & 1.13          & 0.08010          & \third{0.2857}  & -0.5444         \\
        Blender              & 2.07          & 0.07338          & 0.2815          & -0.4427         \\
        Falcor               & 1.53          & 0.15173          & 0.2515          & -0.9577         \\
        UE5                  & 2.40          & 0.07388          & 0.2780          & -0.5214         \\
        \bottomrule
    \end{tabular*}
    \vspace{-0.3cm}
    \label{tab:baseline-real}
\end{table}

\begin{table}[h]
    \centering
    \caption{Quantitative comparsion on PBR synthesis. "$*$" means masked images on single object scenes.}
    \vspace{-10pt}
    \begin{tabular*}{\linewidth}{l c c c c}
        \toprule
        &PSNR $\uparrow$ &  SSIM $\uparrow$ &LPIPS $\downarrow$ & CLIP (Global-I) $\uparrow$  \\
        \midrule
        Ours-$\phi$  & \textbf{30.17} & \textbf{0.9520} &\textbf{0.0469} &\textbf{0.9907}  \\
          RGB\(\leftrightarrow\)X  & 9.6972 &0.3915&0.4986 &0.8861\\
          RGB\(\leftrightarrow\)X*  & 21.61 &0.9129&0.0796& 0.9497\\
        \bottomrule
    \end{tabular*}
    \vspace{-15pt}
    \label{tab:recon}
\end{table}

\begin{table*}[t]
    \centering
    \caption{Quantitative comparison of semantic-aware editing. 
    We evaluate our method against various editing methods. FLUX.1-Kontext-dev serves as the backbone for our dataset construction, while Nano Banana represents a state-of-the-art commercial solution. Our method consistently ranks high across metrics, demonstrating superior overall quality while maintaining a robust balance between background preservation accuracy and semantic alignment.
    \colorbox{best}{Red}, \colorbox{second}{Orange}, and \colorbox{third}{Yellow} indicate 1st, 2nd, and 3rd best results.}
    \vspace{-10pt}
    \label{tab:comparison_merged_final}
    \resizebox{\linewidth}{!}{%
        \begin{tabular}{lccccccccccc}
            \toprule
            & \multicolumn{3}{c}{\textbf{BG Accuracy}} & \multicolumn{4}{c}{\textbf{Semantic Alignment}} & \multicolumn{4}{c}{\textbf{Overall Quality}} \\
            \cmidrule(lr){2-4} \cmidrule(lr){5-8} \cmidrule(lr){9-12}
            \textbf{Method} & PSNR $\uparrow$ & SSIM $\uparrow$ & DINO $\uparrow$ & Global-T $\uparrow$ & Local-T $\uparrow$ & Dir-T $\uparrow$ & BLIP $\uparrow$ & IR $\uparrow$ & Aes $\uparrow$ & $\Delta$CLIP-T $\uparrow$ & Q-Score $\uparrow$ \\
            \midrule
            Ours-$\rho$ & \second{25.46} & \second{0.8641} & \third{0.9428} & \first{0.2875} & \first{0.2458} & \second{0.0887} & \second{0.6762} & \first{-0.2795} & \first{3.9603} & \third{0.0381} & \first{0.3807} \\
            FLUX.1-Fill-dev & \first{31.62} & \first{0.9452} & \first{0.9883} & 0.1919 & 0.2055 & 0.0740 & 0.3798 & -1.3512 & 3.9582 & 0.0264 & \third{0.3366} \\
            FLUX.1-Kontext-dev & 15.39 & 0.4575 & 0.6764 & \second{0.2830} & \second{0.2214} & 0.0748 & \first{0.7486} & \second{-0.3017} & 3.9577 & \first{0.0424} & 0.2877 \\
            DiffEdit & 24.13 & 0.8397 & 0.9276 & 0.1664 & 0.1805 & 0.0002 & 0.0443 & -1.9112 & \third{3.9596} & 0.0009 & 0.2952 \\
            ICEdit & \third{25.12} & \third{0.8440} & 0.8418 & \third{0.2806} & \third{0.2194} & \third{0.0773} & 0.3635 & -0.5022 & \second{3.9601} & \second{0.0392} & \second{0.3470} \\
            Nano Banana & 23.81 & 0.8163 & \second{0.9683} & 0.2783 & 0.1987 & \first{0.1088} & \third{0.6595} & \third{-0.4468} & 3.9592 & 0.0319 & 0.3166 \\
            \bottomrule
        \end{tabular}%
    }
    \vspace{-10pt}
\end{table*}

\vspace{-15pt}
\subsubsection{User Study.}
To quantify the perceptual photorealism of our PRR results, we conducted a user preference study utilizing 15 distinct scenes. We selected representative methods from three primary categories: standard graphics engines, state-of-the-art generative models, and our approach. For physical rendering pipelines, we included high-quality results from industry-standard rendering engines, specifically Unreal Engine 5, Mitsuba3~\cite{Mitsuba3}, Blender, and Falcor~\cite{falcor}. For the generative baseline, we selected FLUX.1-Kontext-dev~\cite{kontext}, NanoBanana~\cite{nanobanana2025} and IntrinsicControlNet~\cite{Lu_2025_ICCV}. Additionally, to validate the incremental efficacy of our approach, we included results generated at different PRR steps of our framework.

The study involved a total of 40 participants, comprising a balanced distribution of genders with ages ranging from 20 to 50. Crucially, the participant pool was composed of both domain experts (including computer graphics researchers and 3D artists) and general users, ensuring that the evaluation captured both professional technical scrutiny and general aesthetic preference. For each trial, participants saw a shuffled set of images and were asked to rank them based on how much they looked like a real photograph. For detailed calculation formulas and other user studies, please see \textit{Appendix D}.

\section{Experiments}

\subsection{Comparison with Baselines}
\label{sec:cp_bs}
We evaluate our proposed framework against state-of-the-art baselines across diverse applications, including PBR image synthesis, image editing, and relighting (with relighting results detailed in the \textit{Appendix A.5}). To ensure a fair and reproducible comparison, we select baselines whose official training or inference codes are available. Given that no single existing framework encompasses this unified capability, we compare against specialized, task-specific methods to evaluate our performance in each application domain.

\subsubsection{Photorealism Enhancement over Graphics Engines.}
Achieving photographic realism in traditional 3D engines demands immense human labor, time, and financial resources to meticulously tune assets, lighting, and materials. To circumvent these prohibitive expenses, our framework automatically transforms raw, unpolished engine renders into highly authentic photorealistic results.

We demonstrate this efficient photorealism upgrade against standard graphics engines, including Mitsuba3~\cite{Mitsuba3}, Blender, Falcor~\cite{falcor}, and UE5, as illustrated in Fig.~\ref{fig:comparison-engine}. While these engines provide physically based foundations, their raw renders often exhibit an idealized synthetic appearance characteristic of pristine surfaces and simplified shading. Our method effortlessly bridges this gap through our PRR process: in the third row, the model introduces realistic weathering effects to building facades and enriches the sparse layout with vehicles and dynamic lighting; similarly, in the second row, it enhances material authenticity by adding organic details such as surface grease and steam. Beyond establishing this foundational realism, our framework further enables diverse photorealistic stylizations, with extended visual results provided in the \textit{Appendix A.1}. 

Quantitatively, this improvement is established by~\refTab{tab:baseline-real}, where our method achieves superior aesthetic and realism scores (e.g., User Study, KID) compared to the raw engine outputs, confirming its capability to generate the authentic textures that standard rasterization typically omits.

\subsubsection{Controlable Image Synthesis over Generative Models.}
Beyond upgrading graphics engines, we evaluate both our PBR and PRR synthesis capabilities against current state-of-the-art generative models. Visual comparisons for both indoor and outdoor environments are presented in Fig.~\ref{fig:comparison-1}, while additional results on single-object assets are provided in the \textit{Appendix A.2}.

For PBR synthesis, we evaluate the physical accuracy by comparing our method with RGB$\leftrightarrow$X~\cite{zeng2024rgb}. As observed in the third row of Fig.~\ref{fig:comparison-1}, our method accurately reconstructs high-frequency details such as the specular reflections on the glossy floor. In contrast, RGB$\leftrightarrow$X introduces significant lighting artifacts, often manifesting as unnatural halos or erroneous shadow rendering rather than faithful global illumination. Notably, even though these indoor scenes are included in the training dataset of RGB$\leftrightarrow$X, it still fails to reproduce the ground truth fidelity. 
Quantitatively, as shown in~\refTab{tab:recon}, standard RGB$\leftrightarrow$X yields poor initial metrics due to its limited generative capability, frequently hallucinating artifacts into the empty black backgrounds of single-object scenes. To ensure fairness, we masked out these background errors (denoted as RGB$\leftrightarrow$X$^*$). Even with this advantage, our method consistently achieves superior accuracy over RGB$\leftrightarrow$X$^*$, confirming our robust capability to produce clean and ground-truth-aligned results.


For PRR synthesis, we compare against IntrinsicControlNet~\cite{Lu_2025_ICCV}. While this method generates realistic textures, it suffers from severe structural degradation. For instance, in the bedroom study depicted in the second row of Fig.~\ref{fig:comparison-1}, IntrinsicControlNet distorts the bookshelf geometry and blurs book details. Furthermore, we also evaluate our generation quality against GAN-based EPE~\cite{EPE}, which frequently yields unnatural and highly unrealistic appearances (detailed visual comparisons are provided in the \textit{Appendix A.3}). By leveraging our P2P training strategy and dataset, our model effectively strikes a superior balance between structural fidelity and authentic, creative generation.

\subsubsection{Progressive Image Editing.}
\label{sec:Subset-Editing}
To comprehensively evaluate the versatility of our framework, we categorize progressive image editing into two distinct paradigms based on user intent. The first focuses on \textit{explicit physical manipulation via subset editing}, where users can either directly alter specific G-buffer channels or use descriptive prompts to achieve precise, decoupled modifications on targeted physical attributes. The second focuses on \textit{semantic photorealistic enhancement via full-channel editing}, where the entire set of G-buffers serves as a unified physical foundation. In this mode, users rely on text prompts and localized masks to directly inject photorealistic details and complex contextual changes into the scene.


Enabled by explicit G-buffer conditioning, our model can accurately execute modifications on target attributes. We demonstrate superiority in this subset G-buffer editing capability compared to baselines: RGB$\leftrightarrow$X~\cite{zeng2024rgb} and IntrinsicEdit~\cite{intrinsicedit}.
We evaluate subset G-buffer editing by applying progressive single-step modifications to distinct G-buffer channels. As observed in Fig.~\ref{fig:ablation_channel}, While baselines capture basic global color shifts, they exhibit significant limitations when handling precise, single-attribute edits. IntrinsicEdit, relying on global prompt embeddings, struggles to interpret fine-grained spatial instructions and explicitly fails to execute localized lighting edits; for instance, when modifying the irradiance highlight, it unintentionally alters the hanging painting and floor textures. Meanwhile, RGB$\leftrightarrow$X struggles to disentangle inputs and align with new conditions; for example, it paradoxically intensifies the wall highlight rather than removing it (top row) and ignores semantic prompts to flatten the quilt (third row). In contrast, our method robustly executes the user's intent in a single step, cleanly applying edits to the targeted physical attributes without ambiguity or artifacts.

Furthermore, this channel-decoupled control successfully extends to multi-step photorealistic relighting. Additional visual cases for both progressive subset editing and relighting are provided in the \textit{Appendix A.4, A.5}.


To evaluate localized photorealism enhancements, we compare models on a progressive semantic-aware editing task (Fig.~\ref{fig:mask_real}). Models must progressively inject photorealistic details via text prompts while confining edits within given semantic masks. Existing baselines struggle to balance localized control and photorealism. Lacking mask constraints, FLUX.1-Kontext-dev~\cite{kontext} alters unmasked backgrounds, causing severe spatial shifts and accumulating noise. Conversely, methods restricting edits via explicit masks (FLUX.1-Fill-dev~\cite{flux2024}) or implicit prompts (Nano Banana~\cite{nanobanana2025}, ICEdit~\cite{zhang2025icedit}, DiffEdit~\cite{couairon2022diffedit}) fail to achieve true realism. FLUX.1-Fill-dev and Nano Banana yield flat CG appearances. Furthermore, ICEdit and DiffEdit suffer severe generative degradation. ICEdit introduces chaotic noise like bizarre scaly wall artifacts, and DiffEdit produces extreme structural distortions that break visual consistency.

In contrast, guided by physical G-buffers and explicit masks, our method strictly confines high-fidelity enhancements (e.g., nuanced lighting, weathering, and objects) to designated areas. Consequently, unmasked regions remain entirely unaltered, ensuring a seamless, artifact-free accumulation of photorealistic details. Quantitatively (\refTab{tab:comparison_merged_final}), our method consistently achieves a superior balance between background preservation and realistic semantic alignment (e.g., top Q-Score and Local-T), effectively overcoming the inherent trade-offs of baselines. Additional progressive editing results are in the \textit{Appendix A.6}.

\begin{figure*}[h]
    \centering
    \includegraphics[width=1\linewidth]{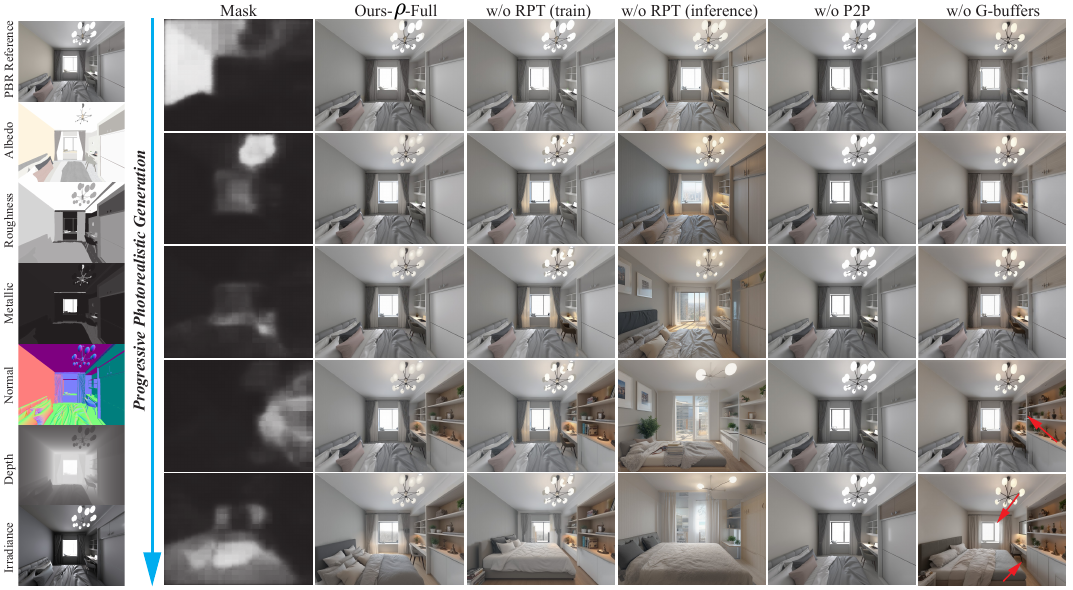}
    \vspace{-20pt}
    \caption{Visual ablation study of our PRR generation method. The generation proceeds progressively from top to bottom, where each row represents a generation stage building upon the result of the previous step and the corresponding G-buffer feature (shown on the left). We compare Ours-$\rho$-Full against variants without residual perceptual transfer mechanism (RPT) during training/inference, without our P2P-50K dataset (P2P), and a direct P2P translation in the RGB domain without G-buffer guidance (w/o G-buffers). The "Mask" column visualizes the semantic injection regions at each stage.}
    \label{fig:ablation_1}
    \vspace{-0.3cm}
\end{figure*}

\begin{figure*}[h]
    \centering
    \includegraphics[width=1\linewidth]{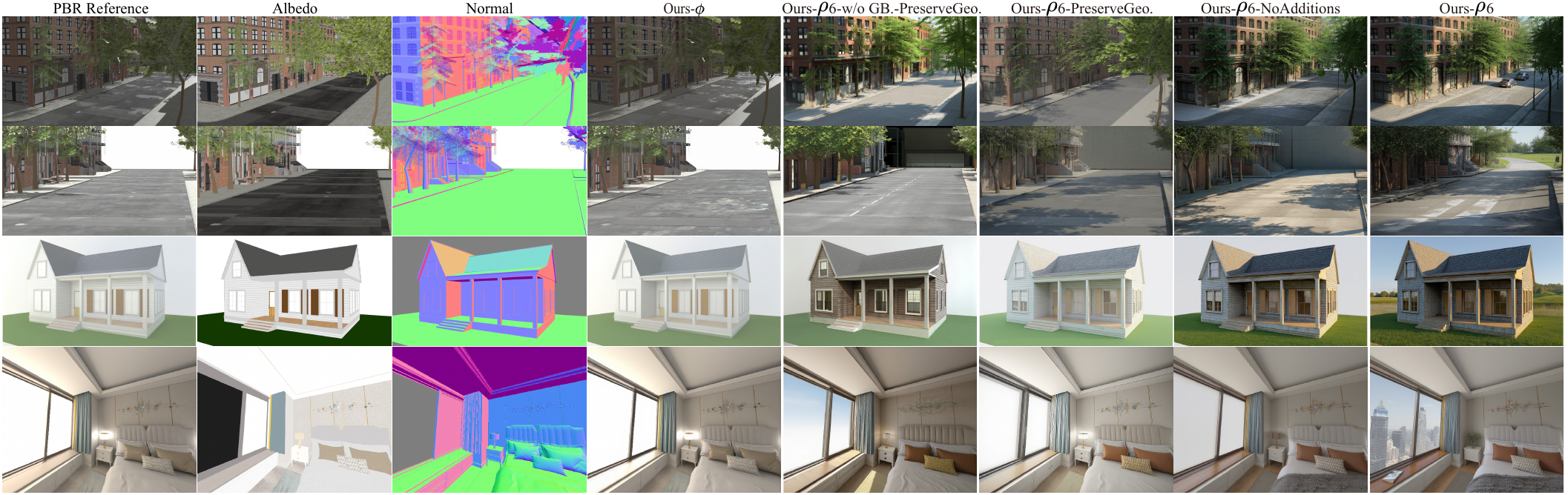}
    \vspace{-20pt}
    \caption{Controllable Photorealism via G-buffer Relaxation. Enabled by our explicit G-buffer controls, users are granted the flexibility to dictate the level of realism by progressively relaxing g-buffer constraints. Notably, explicitly instructing the model to preserve geometry via text prompts without G-buffer guidance (Ours-$\rho$6-w/o GB.-PreserveGeo.) fails to prevent severe structural alterations. While Ours-$\rho6$-PreserveGeo. strictly maintains the original layout, Ours-$\rho6$-NoAdditions enhances materials purely on existing structures without generating new objects. Ultimately, allowing the model to naturally introduce context-aware additions (Ours-$\rho6$) often yields the most authentic and photorealistic results.
}
    \label{fig:gbuffer-necessary}
\end{figure*}

\begin{figure*}[h]
    \centering
    \includegraphics[width=1\linewidth]{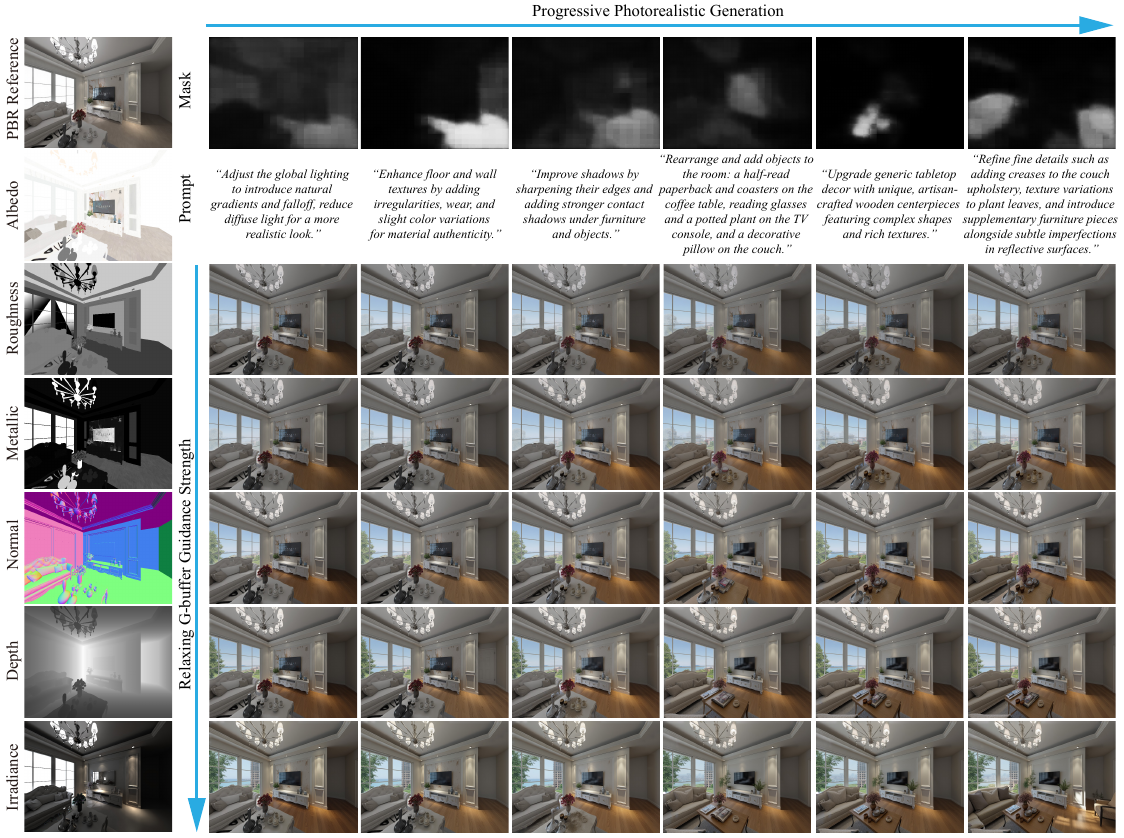}
    \vspace{-20pt}
    \caption{Impact of G-buffer guidance scale ($c\_{scale}$). We visualize the progressive photorealistic generation across varying levels of physical control. For a fair and direct comparison, the exact same textual prompts and regional masks are applied across all different scale settings. As generation progresses (left to right), strong guidance (top rows) enforces strict spatial alignment with the input G-buffers. Conversely, relaxing $c\_{scale}$ (bottom rows) grants the model greater freedom to inject richer textures and complex object variations. This demonstrates that regulating $c\_{scale}$ empowers users to flexibly balance strict structural preservation and generative creativity depending on their specific intents.}
    \label{fig:comparison-cscale-consistency}
\end{figure*}

\definecolor{bg-red}{HTML}{FFCCCC}   
\definecolor{bg-orange}{HTML}{FFE5CC} 

\begin{table}[t]
\centering
\caption{Quantitative ablation study. BG Acc., SA, P2P, RD, RPT, and GB. stand for background accuracy, semantic alignment, our P2P dataset, random dropout, residual perceptual transfer mechanism, and G-buffers, respectively. Photorealism denotes step-average metrics. Results show P2P is vital for realism, while other components mainly remain semantic alignment and background preservation.}
\vspace{-10pt}
\label{tab:ablation}

\resizebox{\columnwidth}{!}{%
\setlength{\tabcolsep}{2.5pt} 
\begin{tabular}{l c c c c c c}
\toprule
 & \multicolumn{2}{c}{\textbf{BG Acc.}} & \textbf{SA} & \multicolumn{3}{c}{\textbf{Photorealism}} \\
\cmidrule(lr){2-3} \cmidrule(lr){4-4} \cmidrule(lr){5-7}
\textbf{Method} & PSNR$\uparrow$ & SSIM$\uparrow$ & CLIP(Loc-T)$\uparrow$ & IR$\uparrow$ & KID$\downarrow$ & CLIP(Glob-T)$\uparrow$ \\
\midrule
Ours-$\rho$ & \cellcolor{bg-orange}25.46 & \cellcolor{bg-orange}0.8641 & \cellcolor{bg-orange}0.2458 & \textbf{-0.2795} & \textbf{0.0569} & \textbf{0.2875} \\
w/o P2P & \cellcolor{bg-red}\textbf{29.79} & \cellcolor{bg-red}\textbf{0.9332} & 0.2326 & -0.5131 & 0.0936 & 0.2779 \\
\midrule
w/o RD & 25.27 & 0.8599 & 0.2430 & -- & -- & -- \\
w/o RPT (tr) & 23.02 & 0.8072 & 0.2453 & -- & -- & -- \\
w/o RPT (inf) & 17.64 & 0.5365 & \cellcolor{bg-red}\textbf{0.2510} & -- & -- & -- \\
w/o GB. & 23.78 & 0.8255 & 0.2421 & -- & -- & -- \\
\bottomrule
\end{tabular}%
}
\vspace{-16pt}
\end{table}

\begin{figure*}[!h]
    \centering
    \includegraphics[width=1\linewidth]{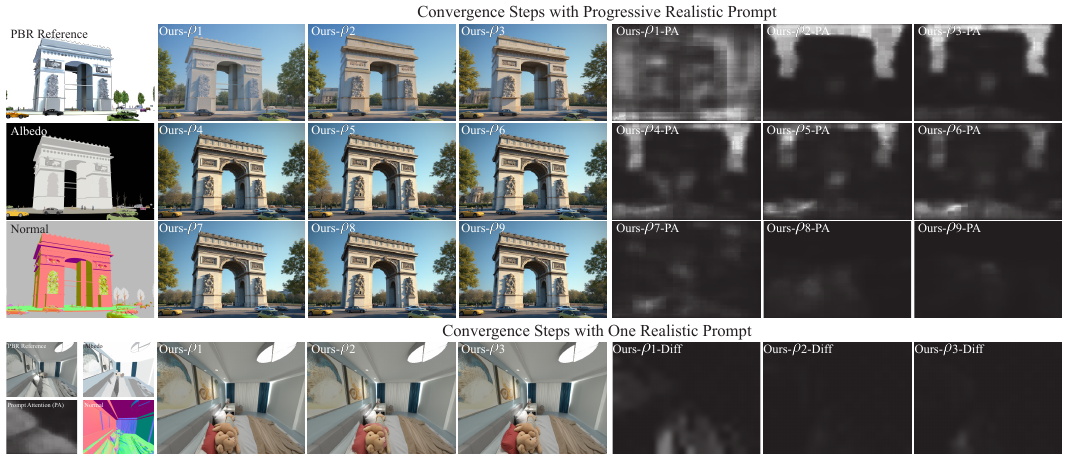}
    \vspace{-20pt}
    \caption{Convergence analysis: progressive and single realistic prompt. The top panel shows progressive generation guided by sequential VLM prompts and their corresponding Prompt Attention (PA) maps. The bottom panel shows generation with a single static prompt, displaying the target PA mask and pixel-wise residuals (Diff) between steps. Overall, our method robustly converges without structural collapse, achieving gradual stabilization under progressive prompts and immediate execution under a single prompt.
}
    \label{fig:converge}
    \vspace{-0.3cm}
\end{figure*}

\subsection{Analysis of G-buffer Guidance}
While~\refSec{sec:Subset-Editing} demonstrates how explicit G-buffers enable decoupled attribute editing, their presence is equally vital during the P2P transition. Specifically, explicitly conditioning on G-buffers provides two critical capabilities that pure RGB-editing methods cannot achieve: (1) preventing unintended scene shifting and structural deformation after continuous multi-step editing, and (2) empowering users to control the exact degree of generated photorealism by selectively applying G-buffer channels and adjusting their guidance strength based on individual intent. We evaluate this indispensability from these two perspectives.

\subsubsection{The Necessity of G-buffer for Spatial Alignment.}
To verify that G-buffers are mandatory for preventing spatial shifts, we compare our full model against a variant stripped of G-buffers (i.e., operating solely in the RGB domain). Removing G-buffer guidance from our pipeline causes the generated structures to explicitly drift from the original layout. As corroborated by the progressive ablation in Fig.~\ref{fig:ablation_1} (rightmost column), the absence of G-buffers leads to accumulating spatial inconsistencies and unintended scaling throughout the continuous multi-step process-specifically, note the progressive drift of the right-side bookshelf, chair, and window, as indicated by the red arrows . In contrast, directly injecting G-buffers guarantees strict spatial alignment over multiple steps. This conclusion is quantitatively supported by \refTab{tab:ablation}, where the ``w/o GB.'' variant exhibits a distinct drop in background accuracy (e.g., PSNR and SSIM) compared to our full method.


\subsubsection{Balancing G-buffer Consistency and Photorealism.} Furthermore, using G-buffers as explicit inputs empowers users to easily control the generated realism based on their specific intent. 
As demonstrated in Fig.~\ref{fig:gbuffer-necessary} (Ours-$\rho$6-PreserveGeo.), strictly maintaining the original geometric layout often yields outputs that look less photorealistic. However, attempting to maintain this layout purely via text prompts without G-buffer guidance (Ours-$\rho$6-w/o GB.-PreserveGeo.) fundamentally fails. Even when explicitly instructed by the prompt to ``preserve geometry,'' the model still suffers from severe spatial shifts and unintended structural alterations; for instance, in the second row, it distorts the shape and structure of the roadside trees. In contrast, explicit G-buffer conditioning provides a robust foundation, allowing users to reliably dictate how strictly to enforce these physical constraints. Beyond rigid adherence, users can choose to enhance materials on existing structures without adding new items (Ours-$\rho$6-NoAdditions), or fully embrace context-aware additions (Ours-$\rho$6) for maximum photorealism.

Furthermore, as visualized in Fig.~\ref{fig:comparison-cscale-consistency}, this balance is continuously tunable via the guidance scale ($c_{scale}$). While high intensity strictly enforces the input layout, progressively relaxing $c_{scale}$ grants the model the necessary freedom to naturally inject richer organic textures and complex objects. Ultimately, G-buffers serve as essential controls, granting users the flexibility to balance strict structural adherence with authentic photorealistic creation through explicit channel selection and intensity adjustment.

\subsection{Ablation Study}
\label{sec:ablation}
We conducted ablation experiments on various components of the framework to validate their effectiveness. Further ablation studies are in \textit{Appendix B}.

\begin{figure}[!h]
    \centering
    \includegraphics[width=0.9\linewidth]{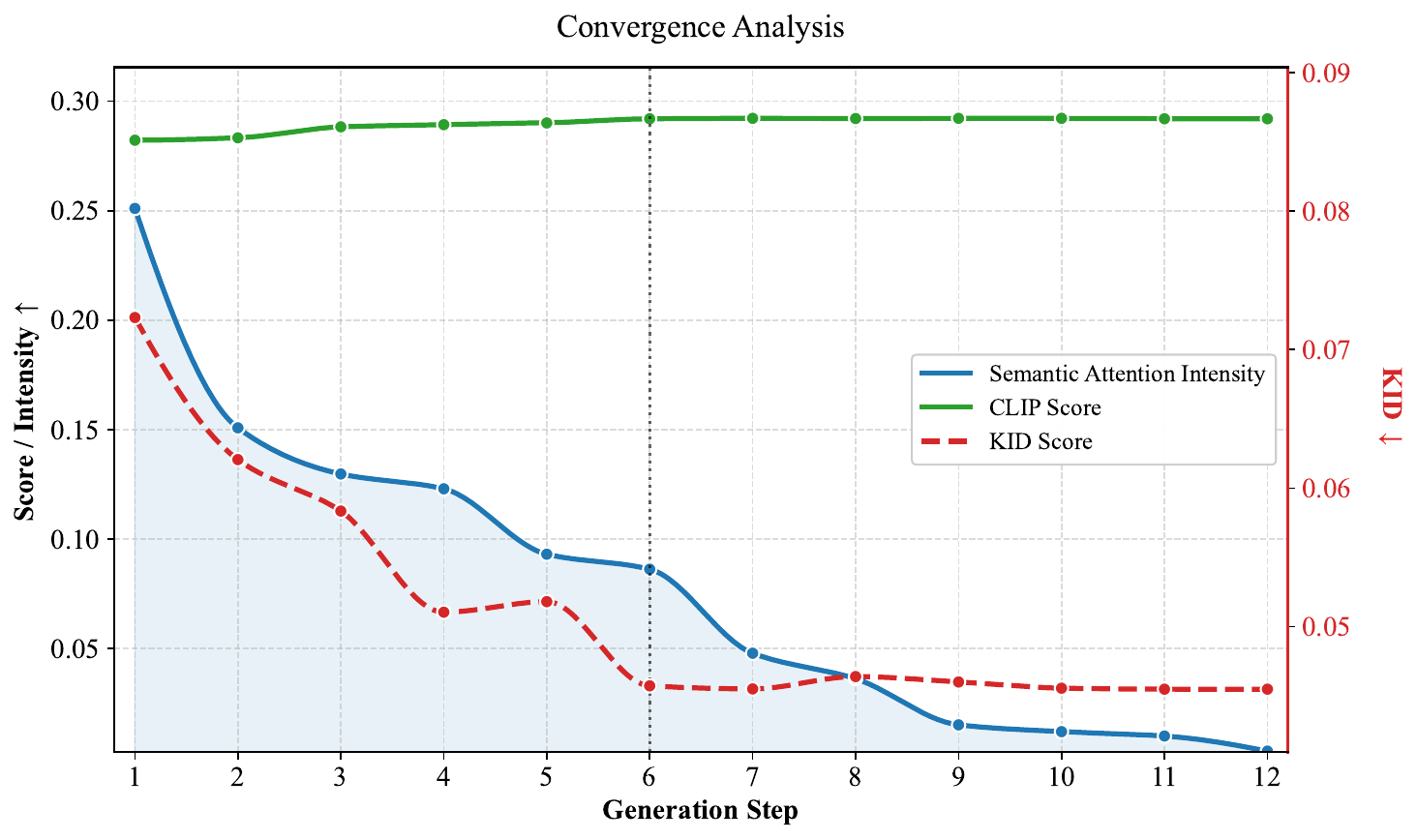}
    \vspace{-15pt}
    \caption{Quantitative convergence analysis under the progressive iterative realistic prompt setting. We monitor the trajectories of Semantic Attention Intensity (blue), CLIP Score (green), and KID Score (red) across 12 generation steps. The vertical dotted line marks all metrics reach convergence at around step 6$\sim$9.}
    \label{fig:converge_analysis}
    \vspace{-0.3cm}
\end{figure}

\begin{figure}[h]
    \centering
    \includegraphics[width=1\linewidth]{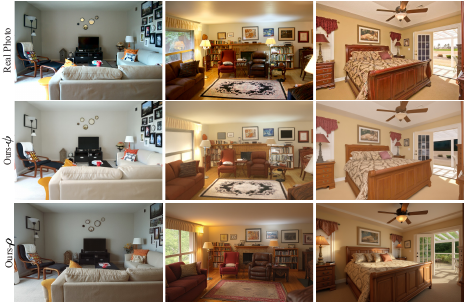}
    \vspace{-20pt}
    \caption{Robustness on inversed G-buffers from real photos. Our method demonstrates strong robustness when applied to G-buffers estimated directly from real-world images. Specifically, Ours-$\phi$ strictly adheres to the physical features of the inversed G-buffers, while Ours-$\rho$ successfully recovers high-fidelity photorealism, achieving a visual quality comparable to the original real photos.}
    \label{fig:real_gbuffer_robust}
\end{figure}

\definecolor{best}{HTML}{FFCCCC}   
\definecolor{second}{HTML}{FFE5CC} 
\definecolor{third}{HTML}{FFFFCC}  

\subsubsection{Impact of Residual Perceptual Transfer for Progressive Semantic Injection.}
We evaluated the importance of our incremental semantic-aware injection by removing it separately during the training and inference phases, as shown in the fourth and fifth columns of \refFig{fig:ablation_1}. During training, it functions as a spatial constraint that explicitly aligns the incremental changes between steps with the corresponding text prompts. Removing it disrupts this alignment, leading to visual inconsistencies.
While the model retains some capacity to complete the image without this training constraint, omitting the strategy during inference results in a severe loss of spatial control. Specifically, the editing effects tend to drift uncontrollably, causing modified areas to extend far beyond the region defined by the prompt-derived mask.
Quantitatively, this loss of control is reflected in \refTab{tab:ablation}, where the `w/o RPT' settings exhibit a marked decline in PSNR and SSIM. This data confirms that without the incremental semantic injection strategy, the model fails to strictly preserve the unedited background, leading to lower accuracy scores.




\subsubsection{Impact of Constructing Pairwise P2P Dataset.}
To validate our dataset construction, we trained a variant solely on the original synthetic data (Engine Render), excluding the P2P samples generated by FLUX.1-Kontext-dev. Although this ablated model correctly adhered to text prompts and target regions, it produced only marginal modifications without a significant improvement in realism, as shown in the sixth column of \refFig{fig:ablation_1}.
As shown in \refTab{tab:ablation}, the model trained without the pairwaise P2P-50K dataset yields significantly lower aesthetic scores (e.g., CLIP-Global and KID), indicating that it remains stuck in the PBR domain. Thus, our P2P dataset acts as an indispensable bridge, effectively closing the domain gap between PBR G-buffers and PRR images.

\subsubsection{Impact of Random Dropout.}
In addition to changing the full set of G-buffers, our model supports flexible editing on partial channel subsets, a capability enabled by our random dropout strategy. This allows users to apply semantic masks to specific channels while preserving the integrity of others, as shown in \refFig{fig:ablation_channel}. The ablation study reveals that excluding the dropout strategy (RD) leads to severe feature entanglement, as the model over-relies on channel correlations. While ``w/o RD'' can generally manage visually dominant traits like albedo and irradiance, it completely fails to disentangle highly coupled physical features. For instance, modifying the quilt's normal unintentionally altered the albedo. Furthermore, the dropout strategy significantly enhances error-correction capabilities, such as smoothing out manual brush strokes during highlight removal in the first row. While the variant without random dropout retains visible artifacts due to its sensitivity to input noise, our full model successfully refines the input to produce a clean result. This is quantitatively reflected in \refTab{tab:ablation}, where the ``w/o RD'' configuration shows degraded G-buffer consistency scores, confirming that random dropout is essential for learning robust, disentangled representations.

\subsubsection{Convergence Analysis of PRR Steps.}
\label{sec:converge_ablation}
To demonstrate that our PRR process reaches convergence and to validate the reliability of the convergence metric ($I_t$) introduced in \refSec{sec:3-6-converge}, we analyze the P2P transfer under two settings, as established in Fig.~\ref{fig:converge} and Fig.~\ref{fig:converge_analysis}.

First, under the progressively iterative prompt setting (top three rows), the expert VLM continuously critiques the current image and provides revision suggestions for the next iteration. Our PRR results (Ours-$\rho i$) become visually stable after step 6, indicating that the incremental realism details are fully integrated. This stabilization is intuitively reflected in the corresponding Prompt Attention (PA) maps: as generation progresses, the attention regions gradually shrink to specific minor details and visually diminish. This visual convergence correlates perfectly with the quantitative metrics in Fig.~\ref{fig:converge_analysis}, where the Semantic Intensity ($I_t$, blue line) shows a sharp decline, and the KID score (red dashed line) plateaus, confirming that the PRR generation with iterative realistic prompts effectively can converge without over-editing.

Second, in the single prompt setting (bottom rows), convergence is rapid. As visualized by the pixel-wise difference maps (Diff), the residuals turn black almost instantly after the first step. This reveals the primary transformation completes almost instantly, with subsequent steps exhibiting only minor pixel-level oscillations. Such immediate stability confirms that our method successfully prevents the structural collapse and noise accumulation typical of the unconstrained FLUX.1-Kontext-dev backbone, ensuring precise editing without quality degradation.

In summary, these results validate both the effectiveness of the $I_t$ metric and the robust convergence capabilities of our framework. Whether steadily accumulating photorealistic details via iterative prompts or executing immediate modifications via a single prompt, the model reliably converges to the target PRR distribution without suffering from indefinite over-editing or structural collapse.

\subsubsection{Robustness on Inversed G-buffers from Real Photos.}
Although our model has never been trained on G-buffers inversed from real-world images, we further evaluate its generalizability and robustness by applying it to such imperfect estimates.
As illustrated in Fig.~\ref{fig:real_gbuffer_robust}, we inverse physical attributes (e.g., Albedo, Roughness, Metallic, Normal, and Irradiance) from original real images. Unlike clean, PBR G-buffers rendered by graphics engines, these inversed inputs naturally exhibit estimation artifacts and lack high-frequency details.

Despite these suboptimal conditions, our method demonstrates strong robustness. Specifically, our intermediate PBR synthesis (Ours-$\phi$) strictly adheres to the structural and material features of the inversed G-buffers, faithfully reflecting the physical inputs exactly as they are. Subsequently, our PRR generation (Ours-$\rho$) successfully compensates for the fine-grained details and complex shading lost during the inverse rendering stage. By injecting rich, realistic textures and atmospheric effects, Ours-$\rho$ successfully recovers high-fidelity photorealism, achieving a visual quality highly comparable to the original real photos. This confirms that our framework is not hindered by imperfect inverse rendering, and can robustly handle in-the-wild applications.

%% file: section/8_limitation.tex
\vspace{-5pt}
\section{Limitations and Future Work}
While our framework demonstrates significant progress in generating photorealistic results from PBR inputs to PRR images, several limitations remain. 
First, inherited from the underlying diffusion properties, the stochastic nature of the generation process can occasionally lead to uncontrolled hallucinations or inconsistent transitions in fine-grained details that deviate from the physical constraints. 
Second, the precise generation and manipulation of complex optical phenomena such as intricate subsurface scattering or caustic lighting effects remain challenging. 
Third, as our method operates within image domain, it lacks a complete 3D scene representation, which limits its ability to reason about global structure and occlusion relationships. 
Fourth, our pipeline currently struggles to synthesize highly authentic human figures, as creating high-quality pairwise human datasets remains a significant challenge for existing generative models (a visual failure case is provided in Appendix E).

Despite these constraints, we believe our approach establishes a novel and effective route for high-quality photorealistic generation. 
Looking ahead, we aim to address these issues by incorporating explicit 3D structural priors as control conditions, which promises to enhance spatial perception and facilitate multi-frame consistency. 
Furthermore, we plan to extend this generative rendering framework to dynamic scenes and integrate specialized identity-preservation modules to accurately synthesize organic subjects like humans.

%% file: section/6_conclusion.tex
\section{Conclusion}
In this paper, we bridge the critical gap between physically-based rendering (PBR) and photorealistic rendering (PRR), denoted as the P2P gap. While PBR ensures mathematical correctness, it often lacks the rich realism of the real world due to simplified digital assets. To bridge this divide, we proposed GeRM, the first multi-modal generative rendering model designed to unify physical accuracy with generative photorealism.
By formulating the P2P transition as a learnable distribution transfer vector (DTV) field, we bypass the need for an explicit realism metric for optimization. To parameterize this vector field, we developed a multi-condition ControlNet that enables the progressive injection of photorealistic details under unified physical and semantic guidance. Furthermore, our novel residual perceptual transfer mechanism for cross-modal alignment resolves the challenge of associating abstract textual prompts with specific physical regions, ensuring precise and controllable incremental updates. These advancements are underpinned by our construction of P2P-50K, an expert-guided pairwise dataset generated via a multi-agent VLM framework, which provides the necessary trajectory supervision in the absence of explicit metrics.
Extensive experiments across PBR synthesis and editing demonstrate that GeRM effectively navigates the continuum between physical constraints and photorealism, establishing a new framework for controllable neural rendering.